\documentclass[ijgi,submit,moreauthors,pdftex,10pt,a4paper]
{Definitions/mdpi} 
\renewcommand{\linenumbers}{}

\usepackage{epstopdf}
\usepackage[caption=false]{subfig}

\usepackage[sort&compress]{natbib}
\usepackage{amsmath}
\usepackage{amsfonts}
\usepackage{amssymb}
\usepackage{amsbsy}
\usepackage{xspace}
\usepackage{amssymb}

\usepackage{newtxmath}
\usepackage{float}
\usepackage{url}
\usepackage{booktabs, makecell}
\usepackage{cuted}
\usepackage{xcolor}
\firstpage{1} 
\pubvolume{xx}
\issuenum{1}
\articlenumber{5}
\pubyear{2018}
\copyrightyear{2018}
\history{Received: date; Accepted: date; Published: date}

\Title{Enhancing crop classification accuracy by synthetic SAR-Optical data generation using deep learning}

\Author{Ali Mirzaei, Hossein Bagheri $^{*}$ and Iman Khosravi }

\address{%
Faculty of Civil Engineering and Transportation, University of Isfahan, Isfahan, Iran;\\alimirzaei1109@trn.ui.ac.ir (A.M.); i.khosravi@cet.ui.ac.ir (I.K.)

$^{*}$ Correspondence: h.bagheri@cet.ui.ac.ir
\\
}

\abstract{\textcolor{red}{This is the pre-acceptance version, to read the final version, please go to ISPRS Int. J. Geo-Inf. on MDPI, URL: \url{https://www.mdpi.com/2220-9964/12/11/450}.} Crop classification using remote sensing data has emerged as a prominent research area in recent decades. Studies have demonstrated that fusing synthetic aperture radar (SAR) and optical images can significantly enhance the accuracy of classification. However, a major challenge in this field is the limited availability of training data, which adversely affects the performance of classifiers. In agricultural regions, the dominant crops typically consist of one or two specific types, while other crops are scarce. Consequently, when collecting training samples to create a map of agricultural products, there is an abundance of samples from the dominant crops, forming the majority classes. Conversely, samples from other crops are scarce, representing the minority classes.
Addressing this issue requires overcoming several challenges and weaknesses associated with traditional data generation methods. These methods have been employed to tackle the imbalanced nature of the training data. Nevertheless, they still face limitations in effectively handling the minority classes. Overall, the issue of inadequate training data, particularly for minority classes, remains a hurdle that traditional methods struggle to overcome. In this research, 
We explore the effectiveness of conditional tabular generative adversarial network (CTGAN) as a synthetic data generation method based on deep learning network, in addressing the challenge of limited training data for minority classes in crop classification using the fusion of SAR-optical data. Our findings demonstrate that the proposed method generates synthetic data with higher quality that can significantly increase the number of samples for minority classes leading to better performance of crop classifiers. For instance, according to the G-mean metric, we observed notable improvements in the performance of the XGBoost classifier up to 5\% for minority classes.
Furthermore, the statistical characteristics of the synthetic data are similar to real data, demonstrating the fidelity of the generated samples. Thus, CTGAN can be employed as a solution for addressing the scarcity of training data for minority classes in crop classification using SAR-Optical data.}


\begin{document}

\section{Introduction} \label{sec.intro}
Cropland classification using remote sensing data has been among the hot and remarkable topics of research in the last two decades. The capability of remotely sensed data acquired by synthetic aperture radar (SAR) and optical sensors has an undeniable role in estimating the area under cultivation and determining the crop yield by preparing a reliable crop map. The fusion of SAR and optical data can greatly help to improve classification accuracy and to achieve more comprehensive information.

The RapidEye satellite is one of the relatively high-resolution optical satellites that has been used in several recent studies for agricultural applications, especially crop mapping \cite{RN19, RN11, RN14, RN15, RN17, RN1}. The spectral bands of RapidEye have been specially designed for applications related to vegetation analysis and have provided more indexing capabilities for extracting crop types\cite{RN15}. On the other hand, the uninhabited aerial vehicle synthetic aperture radar (UAVSAR) radar satellite has also been one of the most widely used radar sensors in the field of crop mapping in the last few years \cite{RN4, RN20, RN7, RN21, RN22, RN18, RN10}. In addition to providing high spatial resolution pixels, this sensor has all four polarizations. Therefore, it is possible to extract coherent and incoherent decomposition parameters related to vegetation and crop types. The fusion of both sensors has shown a high potential in improving the accuracy of crop mapping in different studies \cite{RN8, RN6, RN9}.

Agricultural regions have distinct cultivation patterns, with each region typically characterized by one or two predominant crops, while other crops are less prevalent. Consequently, the availability of training data for all classes, particularly minority crops, is limited. This scarcity poses a challenge for conventional classifiers, as they struggle to accurately differentiate between minority and dominant classes. Consequently, classes with insufficient training samples often experience misclassification.
For instance, previous research \cite{RN23} demonstrated that the Maximum Likelihood and Fully Connected classifiers exhibited poor performance when trained on datasets with insufficient samples compared to datasets with sufficient samples.
In summary, the heavy emphasis of agricultural regions on a small set of major crops leads to an unfair distribution of training data, resulting in insufficient data for minority crops. This imbalance negatively impacts the performance of conventional classifiers, resulting in misclassification of minority classes.

In order to address the challenge of insufficient training data, various methods are employed in the data preprocessing stage. These methods aim to mitigate the impact of limited samples on classifier performance. For instance, the random under-sampling (RUS) method attempts to equalize the influence of low training samples across all classes by randomly removing samples from the majority classes \cite{RN12}. However, this approach carries the risk of discarding potentially valuable and informative samples that could benefit the classifier. Conversely, the random over-sampling (ROS) method aims to artificially increase the sample size of minority classes by duplicating samples. While this technique may help balance the class distribution, it also introduces the possibility of overfitting the classifier due to the generation of redundant, uninformative data \cite{RN12}. It is important to note that both RUS and ROS methods have their limitations and potential drawbacks. RUS may result in sample loss, while ROS can lead to overfitting.
 
To solve such problems, the synthetic minority oversampling technique (SMOTE) was proposed in which, synthetic data is created based on the feature space similarities between minority class samples and using linear interpolation between K existing samples \cite{RN24}. This method has shown good performance in some studies \cite{RN25, RN24, RN26}. However, this algorithm has also weaknesses. Specifically, SMOTE can be divided into two parts. The first part is the strategy of selecting the available samples to be used in the synthetic data generation stage. Since SMOTE considers the importance of all samples of the minority class to be the same, the generated synthetic data is always accompanied by noise. The second part of SMOTE is the linear interpolation strategy for synthetic data generation. This strategy leads to the generation of almost duplicate data, which will cause overfitting in the classifier training process.
Table \ref{tab44} summarizes previous studies that used data generation methods to improve the accuracy of land use and land cover classification.

\begin{table*}[tb]
\centering\footnotesize \caption{Preview of previous studies that used data generation methods to improve the accuracy of classification.}
\begin{tabular}{*4c}
\toprule
\textbf{Author} 
& \textbf{\thead{Sensor}} 
& \textbf{\thead{Study area}}
& \textbf{\thead{Data generation method}}\\
\midrule

Sani et al,  2017 \cite{RN24} 
& Landsat 7 
& Jakarta City & \thead{Variational semi-supervised\\ learning}\\
Douzas et al, 2019 \cite{RN38} 
 & Landsat 8 & \thead{North-western \\ Portugal} & 
SMOTE \\
Fonseca et al, 2021 \cite{RN56}  
&	  \thead{Hyperion, \\ AVIRIS, ROSIS} 	
& \thead{Okavango Delta, Botswana\\ Pavia, northern Italy\\ Kennedy Space Center, Florida\\ Salinas Valley\\ North-western Indiana} &	SMOTE \\
Fonseca et al, 2021 \cite{RN57}  
&	 \thead{Hyperion, \\ AVIRIS, ROSIS}  
& \thead{Okavango Delta, Botswana\\ Pavia, northern Italy\\ Kennedy Space Center, Florida\\ Salinas Valley\\ North-western Indiana}  & SMOTE \\
Hai Ly et al, 2022 \cite{RN55} 
& Landsat 8 
&  Central region of Vietnam & SMOTE \\
Hamid Ebrahimy et al, 2022 \cite{RN58} 
& Sentinel 2  
& Different parts of Iran & \thead{ROS, SMOTE, Adaptive \\ synthetic sampling} \\
\midrule
\end{tabular}\label{tab44}
\end{table*}

Recent advances in deep generative networks have created many possibilities in the field of synthetic data generation. These networks try to learn the probability distribution of real data and produce high-quality synthetic samples. Typically, generative models have illustrated their good performance in the image and text domains, but they have not achieved much success in producing structured (tabular) synthetic data. In recent years, several studies have focused on improving the performance of generative models, especially generative adversarial networks (GAN), on structured data \cite{RN29}. One of the most important challenges often is the non-Gaussian distribution of features in tabular data \cite{RN28}. As a solution, conditional tabular GAN (CTGAN) has been developed to generate synthetic data by considering the distribution of input features. In the architecture of the CTGAN network, a new normalization method is used to overcome non-Gaussian distributions. Thus, CTGAN can be potentially employed for synthetic feature generation in case of insufficient training data for the crop classification task. 
This paper aims to explore the potential of the CTGAN network in addressing the challenge of crop classification using unbalanced tabular samples that comprise optical and SAR polarimetric features. While previous studies have examined the capabilities of the CTGAN network in various data science domains, this study focuses on investigating the network's effectiveness in mitigating the impact of insufficient training data for agricultural product classification using SAR-Optical derived features. 
\textcolor{blue}{The integration of optical and SAR data simultaneously has the potential to yield significant improvements in classification accuracy. By combining the unique strengths of these two data modalities, such as the spectral information from optical data and the structural information from SAR data, we can obtain a more comprehensive understanding of the target objects or land cover classes. This fusion of information enhances the discriminative power of the classification models and enables them to capture a wider range of features and characteristics. Furthermore, this study aims to explore the capabilities of the CTGAN model in generating synthetic data using both optical and SAR images. CTGAN, as a powerful deep learning-based generative model, has shown promise in generating realistic synthetic data that closely resemble the distribution of the original data. By leveraging this capability, we can effectively augment the training dataset with synthetic samples, thereby increasing data diversity and balancing class distributions. This approach has the potential to address the challenge of limited or imbalanced training data, ultimately improving the classification performance. The investigation of simultaneously integrating optical and SAR data, along with the generation of synthetic data using CTGAN, holds tremendous potential for advancing classification tasks, enhancing accuracy, and improving the effectiveness of analyzing complex Earth observation datasets. This combined approach offers a promising avenue to achieve more precise and reliable results, enabling researchers to extract valuable insights from diverse data sources and address challenges such as imbalanced or limited training data. By leveraging the complementary nature of optical and SAR data and harnessing the synthetic data generation capabilities of CTGAN, we can create a comprehensive dataset that captures the unique characteristics of both modalities and enhances the performance of classification models. Ultimately, this research direction has the potential to significantly contribute to the accuracy, robustness, and reliability of classification analyses in the field of earth observation.}

This manuscript consists of several sections; the literature review and the objective of this investigation were introduced earlier. In the following, the study area and the dataset are introduced in Section \ref{study area}. The details of the proposed method and experimental settings are explained in Sections \ref{method} and \ref{setup}, respectively. 
Section \ref{res} presents the results of experiments. Finally, the paper is concluded with a discussion of the achieved results.

\section{Materials and Methods}\label{study area and data set}
\subsection{Data set and study area}\label{study area}
\begin{figure*}[t]
\begin{center}
\includegraphics[width=1\textwidth]{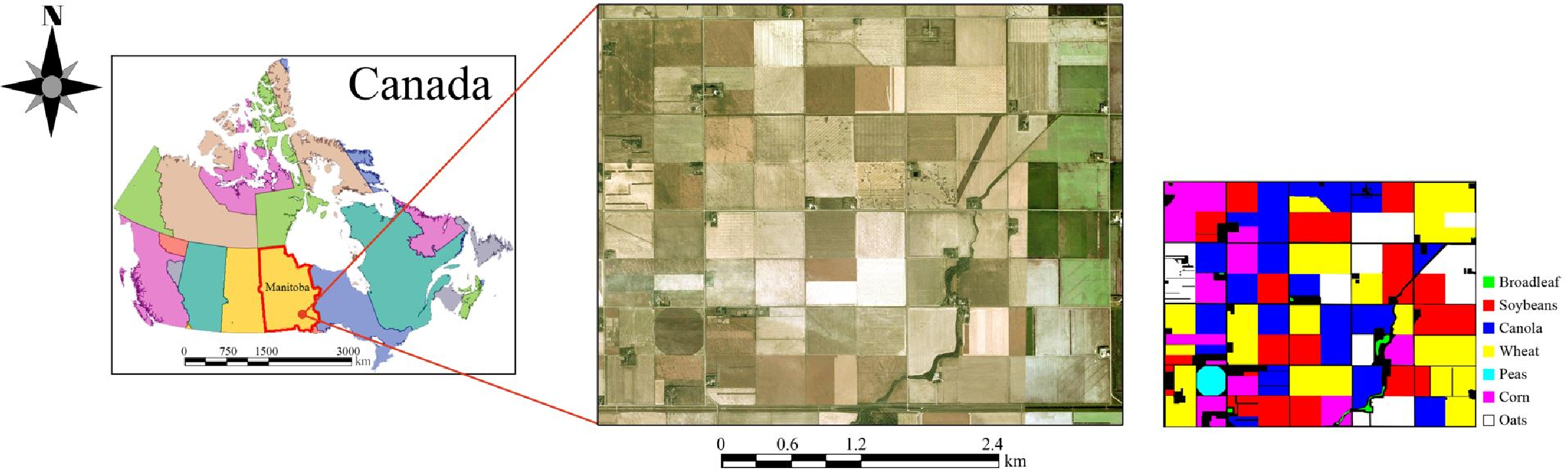}    
\caption{The study area and the reference data in this research.}  
\label{fig1}                                 
\end{center}                                 
\end{figure*}

The study area in this research was an agronomical area of Winnipeg, Manitoba, Canada (see Figure \ref{fig1}). The research used the fused data of bi-temporal optical and PolSAR images. The optical and PolSAR images were acquired from the RapidEye and UAVSAR sensors on 5 and 14 July 2012. Spectral bands of the RapidEye images were blue (B), green (G), red (R), near-infrared (NIR), and red-edge (RE) with a spatial resolution of about 5 m. The UAVSAR images had four polarizations at L-band frequency with a spatial resolution of about 15 m. 

The Soil Moisture Active Passive Validation Experiment 2012 (SMAPVEX 2012) campaign was handled/conducted for the calibration and validation of the National Aeronautics and Space Administration (NASA)’s SMAP satellite over 43 days during the summer of 2012 \cite{RN13}. During this operation, the crop type labels of this data set were collected from the study area including seven classes: broadleaf, canola, corn, oats, peas, soybeans, and wheat. 

Table \ref{tab4} presents the imbalance ratio (IR) in the utilized dataset, which measures the disparity in sample distribution. The IR is calculated as the ratio of $n_{1}$ to $n_{2}$, where $n_{1}$ represents the number of samples in each class and $n_{2}$ represents the number of samples in the majority class. It is evident that there is an imbalanced distribution among the samples across different classes. In particular, the Peas and Broadleaf classes are identified as minority classes, while the remaining classes are categorized as majority classes. 

\begin{table*}[tb]
\centering\footnotesize \caption{The ratio of the number of samples for each class to the number of samples for the majority class.}
\begin{tabular}{*3c}
\toprule
\textbf{Class} 
& \textbf{\thead{Original dataset (IR\%)}} 
& \textbf{\thead{After synthetic data generation (IR\%)}} \\
\midrule

Corn 
& 0.460 
&0.460\\
Peas 
& \textbf{0.002} & 
\textbf{0.12} \\
Canola 
& 0.889 
& 0.889 \\
Soybeans 
&	0.871 	
&	0.871 \\
Oats 
& 0.554 
&	0.554 \\
Wheat 
&	1 
&	1 \\
Broadleaf 
& \textbf{0.002}  
& \textbf{0.12}\\
\midrule
\end{tabular}\label{tab4}
\end{table*}

\subsection{Methodology}\label{method}
\begin{figure*}[t]
\begin{center}
\includegraphics[width=1\textwidth]{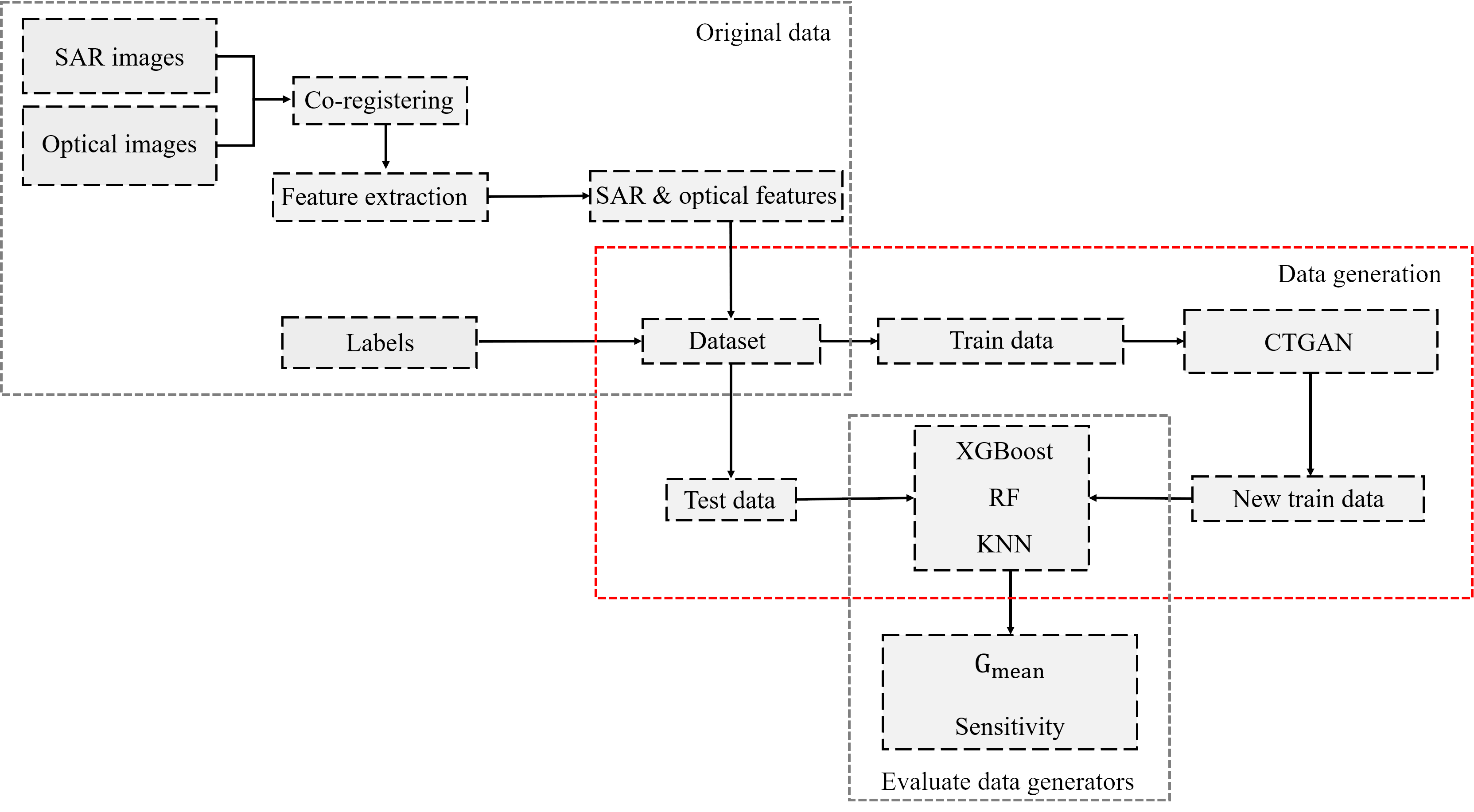}    
\caption{The framework implemented for synthetic data generation and cropland classification.} 
\label{fig2}                                 
\end{center}                                 
\end{figure*}

The detrimental impact of insufficient training samples in the minority classes on classifier performance has been highlighted in the introduction. This study investigates the potential of the CTGAN network in addressing this issue, specifically in the context of agricultural product classification using SAR and optical polarimetric features. The research process is depicted in Figure \ref{fig2}, outlining the key steps involved.
Initially, preprocessing is applied to the SAR and optical images. Before any process, it is necessary to co-register SAR and optical images. These two image sources were co-registered with a linear polynomial for geometrical rectifying and the nearest neighbor method for gray level interpolation \cite{RN54}. Then, various features are extracted from the optical and SAR images in the location of sample data (train and test). More details are explained in the next section. After feature extraction and data preparation, the imbalanced and insufficient training data are fed into CTGAN to rebalance the dataset by generating synthetic samples to increase the sample data of minority classes. The resulting new dataset is then utilized for hyperparameter tuning of different classifiers. Notably, the test samples are separated in advance and are not involved in the process of hyperparameter tuning as well as training the models.
Finally, the quality of the generated synthetic data is evaluated by assessing the performance of the trained classifiers on the independent test data. This evaluation serves to illustrate the effectiveness of the CTGAN network in addressing the challenge of insufficient training data for minority classes in agricultural product classification. More details of each step are described in the following sections.

\subsubsection{Optical and polarimetric feature extraction}\label{features}

In this research, we extracted features from SAR and optical imagery according to the methodology presented in \cite{RN6}. Tables \ref{tab1} and \ref{tab2} present the optical and polarimetric features, respectively. The optical features for RapidEye image included: 5 spectral channels, 17 vegetation indices, and 16 textural indicators which was a total of 38 features. Spectral channels were Blue (B), Green (G), Red (R), Red Edge (RE), and Near Infrared Red (NIR). Vegetation indices were the normalized difference vegetation index (NDVI), simple ratio (SR), enhanced vegetation index (EVI), red-green ratio index (RGRI), atmospherically resistant vegetation index (ARVI), soil adjusted vegetation index (SAVI), normalized difference greenness index (NDGI), green NDVI (gNDVI), modified triangular vegetation index (MTVI2), red-edge normalized difference vegetation index ($NDVI_{re}$), red-edge simple ratio ($SR_{re}$), red-edge normalized difference greenness index ($NDGI_{re}$), red-edge triangular vegetation index ($RTVI_{core}$), red-edge NDVI (RNDVI), transformed chlorophyll absorption in reflectance index (TCARI), triangular vegetation index (TVI) and red-edge ratio 2 (RRI2). 
In addition, eight main parameters of the gray level concurrence matrix (GLCM) of pc1 and pc2, i.e., mean ($\mu$), variance ($\sigma$), homogeneity (HOM), contrast (CON), dissimilarity (DIS), entropy (ENT), angular second moment (ASM), and correlation (COR) were used to describe the textural information. These features provide valuable information about vegetation reflectance in the visible and infrared regions of the electromagnetic spectrum or the spatial characteristics of the crop types \cite{RN16, RN1}.

The polarimetric features for each UAVSAR image included: 6 backscattering intensities, 6 polarization ratios, 6 ratio values, 6 correlation coefficients, 12 Cloud \& Pottier parameters, 2 Pauli parameters, 2 Krogager parameters, 2 Freeman-Durden parameters, and 4 Yamaguchi parameters which were a total of 46 features. It’s noteworthy that $H$, $A$, and $\overline{\alpha}$ are the entropy, anisotropy, and alpha angle, respectively. $\lambda_{1}$, $\lambda_{2}$, and $\lambda_{3}$ are the eigenvalues of the coherency matrix (T), $\psi$ is the pedestal height, and RVI is the radar vegetation index. The polarimetric features give information about the physical and structural properties and also the scattering mechanisms of the various crop types \cite{RN3}.

\begin{table*}[tb]
\centering \footnotesize\caption{The features derived from optical imagery (RapidEye) used in this research.}
\begin{tabular}{*2c}
\toprule
\textbf{Name} & \textbf{Symbol} \\
\midrule
Spectral channels & B, G, R, RE, NIR\\ 
Vegetation indices & \thead{NDVI, SR, RGRI, EVI, ARVI, SAVI, NDGI, gNDVI, MTVI2,\\$NDVI_{re}$, $SR_{re}$, $NDGI_{re}$, $RTVI_{core}$, RNDVI, TCARI, TVI, PRI2}\\
Texture indicators & \thead{$\mu_{pc1}$, $\sigma_{pc1}$, $HOM_{pc1}$, $CON_{pc1}$, $DIS_{pc1}$, $ENT_{pc1}$, $ASM_{pc1}$, $COR_{pc1}$,\\$\mu_{pc2}$, $\sigma_{pc2}$, $HOM_{pc2}$, $CON_{pc2}$, $DIS_{pc2}$, $ENT_{pc2}$, $ASM_{pc2}$, $COR_{pc2}$}\\
\midrule
\end{tabular}\label{tab1}
\end{table*}

\begin{table*}[tb]
\centering \footnotesize\caption{The polarimetric features derived from SAR imagery (UAVSAR) used in this research.}
\begin{tabular}{*2c}
\toprule
\textbf{Name} & \textbf{Symbol} \\
\midrule
Backscattering intensities (dB) & $\sigma_{hh}$, $\sigma_{hv}$, $\sigma_{vv}$, $\sigma_{rr}$,$\sigma_{rl}$, $\sigma_{ll}$\\ 
Polarization ratio (dB) & $R_{hhvv}$, $R_{hvhh}$, $R_{hvvv}$, $R_{rrll}$, $R_{rlrr}$, $R_{rlll}$\\
Ratio values & $R_{hh}$, $R_{hv}$, $R_{vv}$, $R_{rr}$, $R_{rl}$, $R_{ll}$\\
\thead{PolSAR correlation coefficients} & $\rho_{hhvv}$, $\rho_{hvhh}$, $\rho_{hvvv}$, $\rho_{rrll}$, $\rho_{rlrr}$, $\rho_{rlll}$\\
Cloud \& Pottier parameters & $H$, $A$, $\overline{\alpha}$ $HA$, $H$(1–$A$), (1–$H$)$A$, (1–$H$)(1–$A$), $\lambda_{1}$, $\lambda_{2}$, and $\lambda_{3}$, $\psi$, RVI\\
Pauli parameters & $\mid \alpha \mid ^ {2}$, $\mid \beta \mid ^ {2}$ \\ 
Krogager parameters & $\mid k_{d} \mid ^ {2}$, $\mid k_{h} \mid ^ {2}$\\
Freeman-Durden parameters & $P_{s}$, $P_{d}$, $P_{v}$\\ 
Yamaguchi parameters & $Y_{s}$, $Y_{d}$, $Y_{v}$, $Y_{c}$\\
\midrule
\end{tabular}\label{tab2}
\end{table*}

\subsubsection{Machine learning classifiers}\label{3.1}

Classifiers based on machine learning have received much attention in past studies, especially in the field of remote sensing \cite{RN30, RN47, RN48, RN50}. These algorithms have a good ability to model complex classes and understand different input features. Also, they don't need any initial assumptions about data distribution. In general, these algorithms are more accurate than traditional parametric methods, especially in the face of high dimensional data \cite{RN31}.
In this research, three algorithms, i.e., random forest (RF), extreme gradient boosting (XGBoost), and K nearest neighbor (KNN), are used to investigate the performance of the CTGAN network in generating the synthetic SAR-optical features.

RF utilizes the bagging method for training, where base learners (decision trees) are trained independently. In this approach, random sampling with replacement is performed, meaning that data points are randomly selected from the training set. Consequently, a training sample may be selected multiple times within this chosen data. The majority vote of each decision tree's output determines the final output class. By aggregating decision trees, RF is robust against overfitting, capable of identifying outliers, and can assess the importance of input variables. However, by raising the number and complexity of trees, the training and prediction time of the model also increases \cite{RN32}.

XGBoost algorithm, based on gradient-boosted decision trees (GBDT), is another popular machine learning algorithm. It leverages the errors from previous iterations and enhances the importance and weight of incorrectly predicted instances in subsequent iterations. XGBoost incorporates Regularization in the cost function to avoid overfitting and employs parallel processing during training, resulting in faster processing and improved accuracy. Each tree in this algorithm generates an output based on different independent variables. After constructing the trees, the majority of predicted classes determine the class of the input data \cite{RN33}.

KNN is a lazy learning algorithm that operates based on nearest neighbors. It calculates the distance between the test sample and all training samples, selects the K closest samples based on distance, and determines the dominant class among these K samples as the class of the test sample \cite{RN34}.

\subsubsection{Synthetic data generation}\label{3.2}

The introduction highlighted the negative impact of insufficient training samples on the performance of various classifiers. This issue leads to reduced effectiveness of minority classes in minimizing the Loss function during training, resulting in classifier bias towards majority classes \cite{RN35}. To address this challenge, different methods are employed, including random sampling techniques such as ROS and RUS. \textcolor{blue} {The strengths and weaknesses of these methods were already discussed in Introduction.}
Another popular method for generating synthetic data for minority classes is SMOTE. This technique generates artificial samples along the connecting line between K real samples within the minority classes. Previous studies have successfully utilized SMOTE for synthetic data generation \cite{RN24,RN38, RN53}.
However, this method still suffers from some problems such as generating outlier data. Moreover, there are scenarios that using linear interpolation in the SOMTE method will cause the generation of duplicate data and the occurrence of problems such as overlearning \cite{RN38}.

\begin{figure}
\begin{center}
\includegraphics[height=4cm]{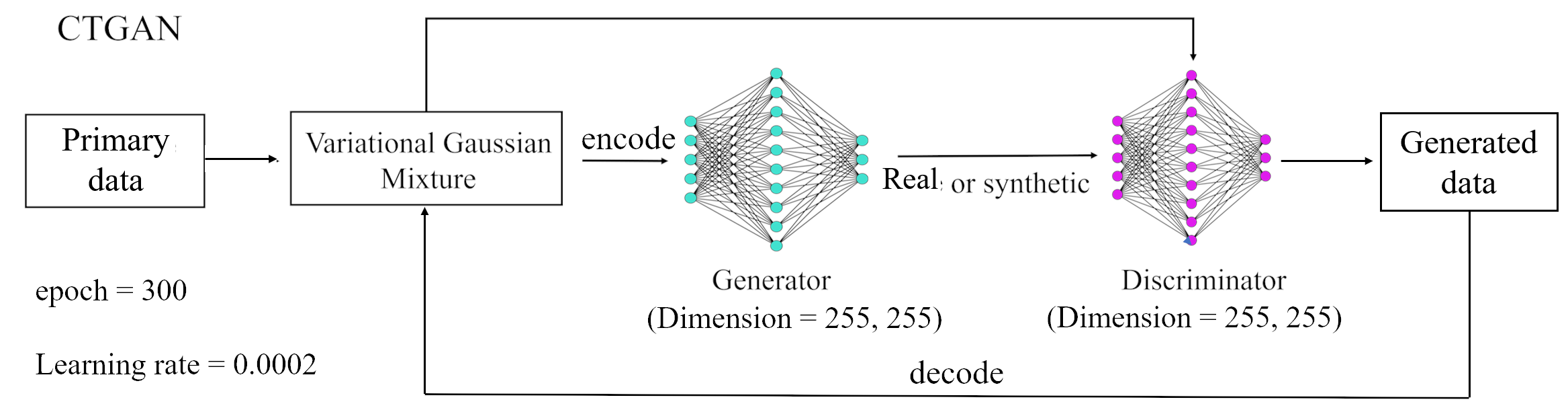}    
\caption{The structure of CTGAN for synthetic data generation.} 
\label{fig3}                                 
\end{center}                                 
\end{figure}

Recently, GAN networks have been employed for generating various types of synthetic data. GAN networks include two parts, a generator, and a discriminator, which learn the distribution of data through an adversarial training process. The task of the generator is to generate synthetic data assuming a Gaussian distribution for the real data. The discriminator is responsible for distinguishing synthetic data produced by the generator from real data. When the generator can defeat the discriminator, the GAN network will be able to produce synthetic data similar to the real data. However, these networks encounter challenges in generating desirable synthetic samples in the case of the non-Gaussian distribution of tabular datasets. To diminish this weakness, the CTGAN network has been proposed.  CTGAN is a special type of GANs designed to generate synthetic structured (tabular) data. Despite the architectural similarities of CTGAN with other GANs, there are also key differences between these networks. First, CTGAN is specifically designed for generating synthetic tabular data, which includes both continuous and categorical variables organized in a tabular format. Second, CTGAN supports conditional data generation, which means users can specify conditions to generate synthetic data of a particular variable. Thirdly, CTGAN uses categorical embeddings to represent categorical variables in the generated synthetic data. This allows CTGAN to effectively handle discrete categorical variables in tabular data. Other GANs may not have specific mechanisms to handle categorical variables or may require additional preprocessing or encoding techniques \cite{RN40, RN28, RN42, RN43}.
\textcolor{blue}{Figure \ref{fig3}  illustrates the overall structure of CTGAN, which includes novel preprocessing techniques to improve GAN performance on tabular data generation. Tabular data distributions may be non-Gaussian. This can cause GANs to struggle with the "vanishing gradient" problem during training. To address this, CTGAN first estimates the underlying distribution of each feature using a Variational Gaussian Mixture Model (VGMM). VGMM represents the overall distribution as a weighted combination of multiple Gaussian components, each with its own mean and covariance. This models multi-modal distributions more flexibly than a single Gaussian. The estimated VGMM distribution is then used to normalize each feature through "encoding". The encoded data has a standardized distribution that helps GAN training converge. The generator produces synthetic samples in this normalized space. After training, a "decoding" step transforms the generated data back to the original distribution through the reverse of the encoding transformations. This preprocessing allows CTGAN to handle complex, non-Gaussian tabular datasets while stabilizing GAN training. The end-to-end framework can generate high-quality synthetic samples in the native distribution of the real data \cite{RN40, RN28}.}

CTGAN network training is based on the following loss function:
\begin{equation}\label{eq.1}\begin{split}
L_{D} = \frac{1}{m} \sum _{i=1}^{m}[D(x^{\prime ^ i})-D(x^{i})]
\\L_{G} = -\frac{1}{m} \sum _{i=1}^{m} [D(x^{\prime ^ i})]+H
\end{split}
\end{equation}
where $L_{D}$  and $L_{G}$ are the loss functions for the discriminator and the generators, respectively, $D(x)$ is the output of the discriminator for real data, $D(x^\prime)$ is the output of the discriminator for synthetic data, $H$ is the cross-entropy score and $m$ denotes the number of synthetic samples \cite{RN40, RN28}.

\subsection{Experimental setups}\label{setup}

During the training process of all algorithms, 10\% of the dataset was allocated as the training dataset, while the remaining data served as the test dataset. 
\textcolor{blue}{To determine the optimal hyperparameters for each classifier, we employed the random search algorithm combined with K-Fold cross-validation strategy. Specifically, we set the value of K to 3, indicating that the training dataset was divided into three subsets (folds) of approximately equal size. During the random search process, different combinations of hyperparameters were randomly sampled from predefined ranges for each classifier. These hyperparameters included parameters such as learning rate, regularization strength, number of hidden layers, and activation functions, among others, depending on the specific classifier being tuned. For each sampled combination of hyperparameters, the classifier was trained on two folds of the training dataset and evaluated on the remaining fold. This process was repeated three times, with each fold serving as the evaluation set once. The evaluation results from the three folds were then averaged to obtain a more robust estimate of the classifier's performance for that particular set of hyperparameters.}
The generator and discriminator components of the CTGAN network were defined using residual fully connected neural networks and linear networks, respectively. The discriminator consisted of two layers, each containing 255 neurons. The CTGAN model was trained in 300 iterations, utilizing an Adam optimizer with a learning rate of 0.002. For comparison purposes, the SMOTE, ROS, and RUS methods were also implemented.

Table \ref{tab40} summarizes the hyperparameters tuned for different classifiers as well as synthetic data generators implemented in this study.
\begin{table*}[tb]
\centering\footnotesize \caption{The hyperparameters tuned for different classifiers and synthetic data generators.}
\begin{tabular}{*3c}
\toprule
\textbf{ \thead{Classifier\\ or \\Data generator}} 
& \textbf{\thead{hyperparameters}} 
\\
\midrule

XGBoost 
& No. of estimators $=$ 1000, max depth $=$ 18, max features $=$ 1, bootstrap $=$ True, max samples $=$ 0.9 
\\
RF 
& No. of estimators $=$1000, max depth $=$ 18, max features $=$ 0.6, bootstrap $=$ True, max samples$=$ 0.8 & 
\\
KNN 
& No. of neighbors$=$ 10 ,weights $=$ distance, metric$=$ Manhattan 
 \\
CTGAN 
&	layers = 2, neurons in each layer=255, Optimizer $=$ Adam,\ learning rate $=$ 0.002, epoch = 300 	
 \\
SMOTE 
& Sampling strategy$=$ not majority, k neighbors$=$ 5 
 \\

\midrule
\end{tabular}\label{tab40}
\end{table*}

According to Table \ref{tab4}, the generated dataset significantly increases the ratio of data imbalance for the majority class compared to the original dataset. Specifically, the ratio after synthetic data generation is 50 times greater than that of the original dataset. (from 0.002 to 0.12). Synthetic samples for the minority classes (pea and broadleaf) were generated using each of the algorithms (CTGAN, SMOTE, and ROS). Additionally, in the RUS method, the number of samples for all classes was reduced to match the number of samples in the minority classes.

In this study, to evaluate the performances of the various data generators, $G_{mean}$ and Sensitivity (recall) were defined as below:

\begin{equation}\label{eq.2}
G_{mean} = \sqrt{Sensitivity \times Specificity}
\end{equation}
\begin{equation}\label{eq.3}
Sensitivity  = \frac{TP}{TP+FN}
\end{equation}
\begin{equation}\label{eq.4}
Specificity = \frac{TN}{TN+FP}
\end{equation}
where TP represents the positive samples that have been predicted as true. FN denotes the negative samples that have been predicted false. TN represents negative samples that have been predicted as true and FP identifies positive samples that have been predicted as false. These values are calculated separately for each class.
According to equations \ref{eq.2} to \ref{eq.4}, 
\textcolor{blue}{The sensitivity metric, also known as the true positive rate or recall, measures the proportion of actual positive samples correctly classified as positive by a classifier. It focuses on correctly identifying samples belonging to the positive class and is particularly sensitive to the classification of a sample in the wrong class. Sensitivity is an important metric, especially in scenarios where the accurate detection of positive instances is critical. On the other hand, the G-mean, or geometric mean, is a metric that evaluates the overall performance of a classifier by considering both the majority and minority classes. It takes into account both the sensitivity (true positive rate) and specificity (true negative rate) metrics. The G-mean is calculated as the square root of the product of sensitivity and specificity, providing a balanced measure of classifier performance across different classes. The G-mean is advantageous when dealing with imbalanced datasets, where the number of samples in one class is significantly smaller than the other. In such cases, accuracy alone can be misleading since a high accuracy can be achieved by simply classifying all samples into the majority class. The G-mean helps to capture the classifier's ability to perform well in both the majority and minority classes, as it considers the trade-off between sensitivity and specificity.
By using the G-mean metric, researchers and practitioners can obtain a more comprehensive evaluation of classifier performance, especially in imbalanced datasets. It provides insights into how well the classifier can handle both positive and negative instances, enabling a more accurate assessment of its effectiveness in real-world applications.
In summary, while sensitivity focuses on the correct classification of positive samples, the G-mean takes into account the performance of classifiers in both majority and minority classes. Together, these metrics provide a more comprehensive understanding of classifier performance and are particularly useful when dealing with imbalanced datasets \cite{RN44}.}
\section{Result}\label{res}
This section presents the performance of implemented synthetic data generation methods for different classifiers. As shown in Table \ref{tab5}, based on the Sensitivity metric, the performance of all three classifiers in detecting minority classes (Peas and Broadleaf) improved after synthetic data generation. The best performance among different methods belongs to CTGAN while maintaining the overall performance of the classifier based on the total Sensitivity metric. The improvement in the XGBoost classifier was 9.4\% and 8.9\% for CTGAN for Peas and Broadleaf classes, respectively, compared to the original dataset. The results of the RF classifier trained and tested on the CTGAN dataset show that the Sensitivity increases to 95\% from 92.8\% and 80\% for both Peas and Boardleaf classes, respectively. Unlike the previous two classifiers, the KNN algorithm performs very poorly in classifying these two classes with imbalanced and insufficient datasets, so that it almost cannot classify any of the samples of these two classes. After data generation by CTGAN, the performance of the KNN algorithm for the Peas class reaches 92.2\%, which is 20.5\%, 57.2\%, and 20.0\% better than the dataset produced by SMOTE, ROS, and RUS, respectively. But for the Broadleaf class, the performance of SMOTE is 2.2\% better than CTGAN.
\textcolor{blue}{Also, according to the obtained results, the performance of RUS is very good in increasing the performance of minority classes based on the Sensitivity metric, but the overall performance of the classifiers demonstrates that the RUS method reduces the total Sensitivity of crop classification.}

\begin{figure}
\begin{center}
\includegraphics[height=22cm]{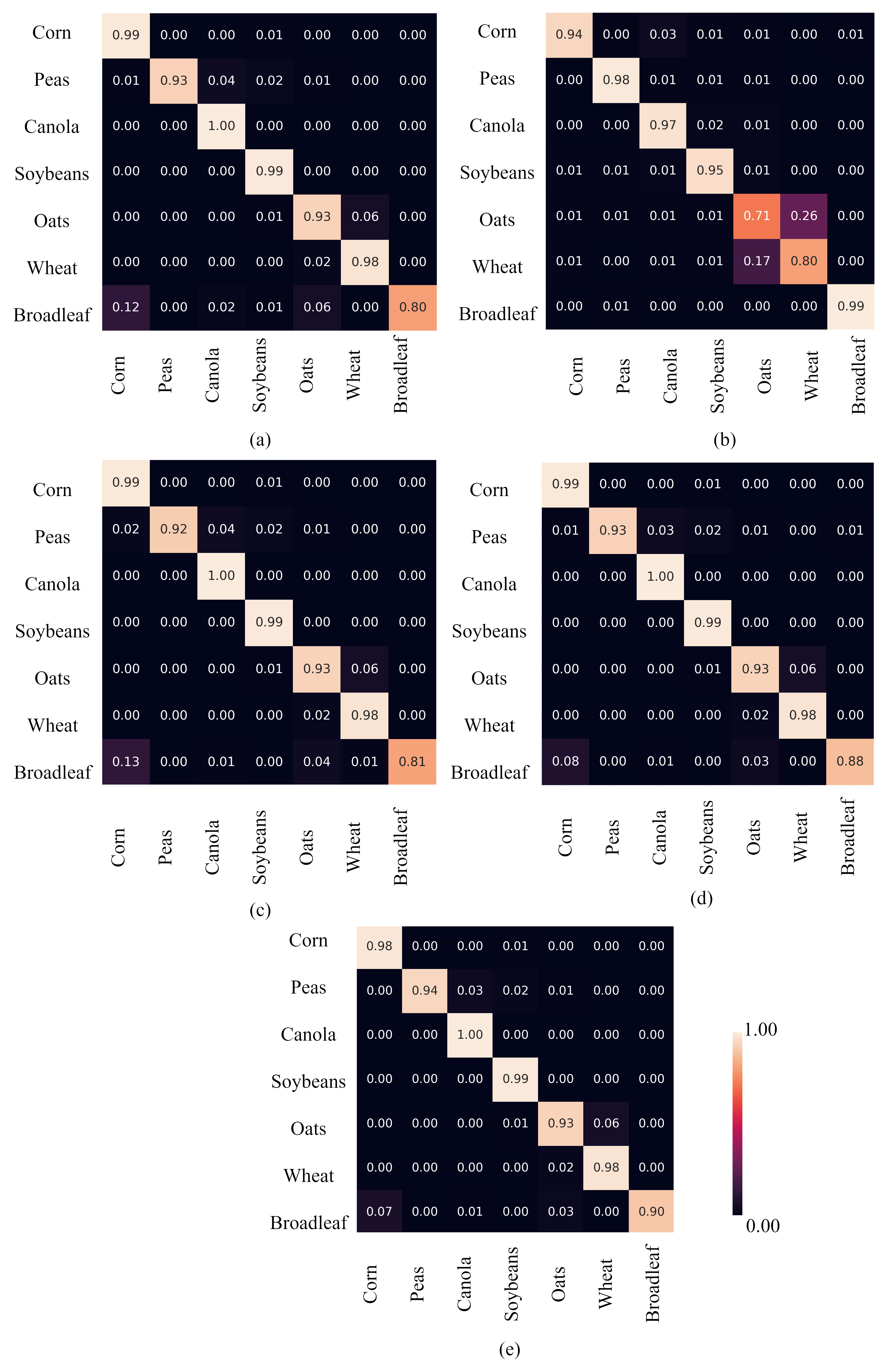}    
\caption{Confusion matrices for the RF classifier trained and tested on: a) Orginal b) RUS c) ROS d) SMOTE e) CTGAN datasets.}  
\label{fig4}                                 
\end{center}                                 
\end{figure}

 \textcolor{blue}{For better evaluation, the confusion matrices of the RF classifier for the original (imbalanced and insufficient), RUS, ROS, SMOTE, and CTGAN datasets are displayed in Figure \ref{fig4}.} Based on this figure, the correctly classified samples for the Peas class in the original dataset and SMOTE dataset are equal to 93\% while this value is 94\% for the CTGAN dataset. In addition, for the Broadleaf class, the amount of correctly classified samples increased from 80\% for the original dataset to 90\% for the  CTGAN datasets, respectively. Despite the balancing using RUS, ROS, and SMOTE, increases the Sensitivity of the minority classes, the performance of the classifier decreases for other classes.

\begin{table*}[tb]
\centering \footnotesize\caption{The sensitivity of different classifiers after data generation with various methods. The sensitivity has been improved for all classifiers after generating data using different methods.}
\begin{tabular}{*{10}{c}}
\toprule
{Classifier} & \multicolumn{3}{c}{XGBoost} & \multicolumn{3}{c}{RF} & \multicolumn{3}{c}{KNN}\\
\midrule
Data set $\setminus$ Class  & Peas & Broadleaf & \thead{Total\\Sensitivity} & Peas & Broadleaf  & \thead{Total\\Sensitivity} & Peas & Broadleaf  & \thead{Total\\Sensitivity} \\
\midrule
Original   & 0.856    & 0.850 & 0.98  & 0.928 & 0.800 & 0.98 & 0.011 & 0.028 & 0.86\\
RUS & 0.950   & 0.983 & 0.87 & 0.983 & 0.989 & 0.88 & 0.722 & 0.583 & 0.49\\
ROS & 0.872   & 0.878 & 0.98 & 0.918  & 0.811 & 0.98  & 0.350  & 0.274 & 0.86\\
SMOTE  &  0.867   & 0.933 & 0.98  & 0.928 &  0.878 & 0.98  & 0.717  & 0.794 & 0.86 \\
CTGAN  &  0.950   & 0.939 & 0.98 & 0.950 &  0.950  & 0.98  & 0.922  & 0.772 & 0.86 \\
\midrule
\end{tabular}\label{tab5}
\end{table*}

The evaluation of crop classification with the XGBoost algorithm using synthetic data generated by RUS, ROS, SMOTE, and CTGAN based on the $G_{mean}$  metric is presented in Table \ref{tab6}. As shown, the classification accuracy is improved for minority classes. The performance improvement for the Peas class is 5.0\% for CTGAN datasets. In addition, in the Broadleaf class, the $G_{mean}$  metric is improved from 92.2\% for the original dataset to 0.969\% for the CTGAN dataset. Also, the performance of the classifiers using the ROS dataset decreased significantly. 

In summary, CTGAN more effectively generated datasets for classes with insufficient samples, while maximizing both overall and minority class performance across classifier metrics, outperforming alternative techniques.

\begin{table*}[tb]
\centering \footnotesize\caption{$G_{mean}$ of XGBoost for different crop classes.}
\begin{tabular}{*6c}
\toprule
Class  & Insufficient and imbalance & RUS & ROS & SMOTE & CTGAN \\
\midrule
Corn   & 0.994 & 0.956 & 0.954 & 0.993 & 0.994 \\
Peas   & \textbf{0.925} & \textbf{0.972} & \textbf{0.591} & \textbf{0.931} &  \textbf{0.975}\\
Canola   &  0.998 & 0.989 & 0.981 & 0.998 & 0.998\\
Soybeans   & 0.995 & 0.950 & 0.947 & 0.994 & 0.995\\
Oats    & 0.973  & 0.849 & 0.787 & 0.970 & 0.973\\
Wheat   &  0.986 & 0.848 & 0.865 & 0.985 & 0.986\\
Broadleaf    & \textbf{0.922} & \textbf{0.990} & \textbf{0.567} & \textbf{0.966} & \textbf{0.969} \\
\midrule
\end{tabular}\label{tab6}
\end{table*}

\section{Discussion}\label{dis}

\begin{figure*}[t!]
\begin{center}
\includegraphics[width=0.6\textwidth]{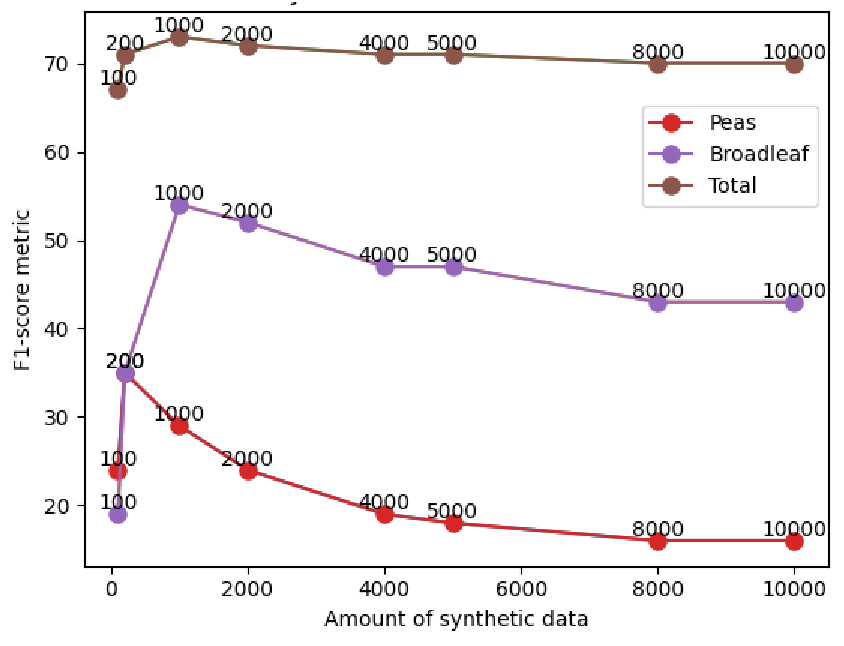}    
\caption{The performance of the KNN classifier, measured by the F1
score metric, in relation to varying quantities of synthetic samples generated by the CTGAN model.}  
\label{fig10}                                 
\end{center}                                 
\end{figure*}

This study aimed to address the issue of limited training samples in minority crop classes by utilizing synthetic data generated by the CTGAN network. Specifically, the proposed approach utilized the fusion of optical and polarimetric SAR features for crop classification. As illustrated in section \ref{res}, the KNN performance was improved significantly by employing the synthetic data generated by CTGAN.
Figure \ref{fig10} demonstrates the KNN classifier performance for different quantities of synthetic samples from 100 to 1000 produced by CTGAN. The red line plots the accuracy of the Peas class. Similarly, the blue and brown lines, respectively depict the overall accuracy based on the F1-score (a comprehensive metric accounting for precision and sensitivity). Precision refers to the proportion of correctly identified positive cases out of all classified as positive.
For this problem, 1000 synthetic samples achieved slightly higher accuracy than other volumes for the classifier and Broadleaf class. The Peas class accuracy peaked at 200 samples. However, the addition of 200 samples did not sufficiently reduce the imbalance in the dataset. Therefore, 1000 synthetic samples were generated for the minority classes, resulting in a 50x reduction in the class imbalance ratio.

While the primary goal of this research was to generate additional data by CTGAN for minority classes in order to address the problem of insufficient training samples, using this data generation method also helped tangentially reduce the class imbalance in the dataset. By increasing the number of samples for minority classes, the technique brought the class distribution closer to a balanced ratio, even though balancing the dataset was not the main focus. Thus, the data generation by the CTGAN approach served a dual purpose; producing more training samples for insufficient classes and mitigating the existing skew between majority and minority classes.

Figure \ref{fig10} also illustrates the ability of CTGAN to produce diverse data volumes, while the optimal should be determined depending on the problem. In summary, CTGAN-generated synthetic data leveraging multimodal crop data helped to boost classifier performance on minority classes. The analysis determined generating 1000 samples per class would achieve a good balance between accuracy and balancing class representation. In summary, the result illustrated the efficiency of CTGAN in addressing limited training data challenges for crop classification tasks.

\subsection{Influence of synthetic data on the performance of classifiers }

The results in Tables \ref{tab5}, \ref{tab6}, and Figure \ref{fig4} show that synthetic data generation can impact the performance of classification depending on the classifier model. For example, KNN benefited more significantly from additional training data compared to RF and XGBoost, whose performance increased to a lesser extent.
Based on G-mean, CTGAN produced higher quality synthetic data that led to greater classification accuracy improvements for minority classes over other methods. However, based on sensitivity, RUS outperformed CTGAN for minority classes, but RUS reduced overall performance by removing useful information from other classes.
While ROS yielded a slight boost to classifiers, its performance was weaker than SMOTE and CTGAN due to providing less new information for training. On the other hand, SMOTE generated synthetic data without considering real data distributions, diminishing accuracy gains shown in Table \ref{tab5}.
CTGAN uniquely can produce a balanced dataset that accurately reflects the real data distribution. This preserves classification performance for the majority class while substantially improving the classification accuracy of the minority classes. Whereas other methods either overfit certain classes or remove informative samples, CTGAN's distribution-aware generation approach leads to well-balanced classification across all classes.

In summary, CTGAN yielded the most robust accuracy improvements by introducing informative synthetic samples without distorting real data properties or removing important information. Its ability to balance datasets while maintaining fidelity to underlying distributions provides an advantage over other data augmentation methods.
\subsection{Quality of generated synthetic data }

\begin{figure}[ht!]
\begin{center}
\includegraphics[height=22cm]{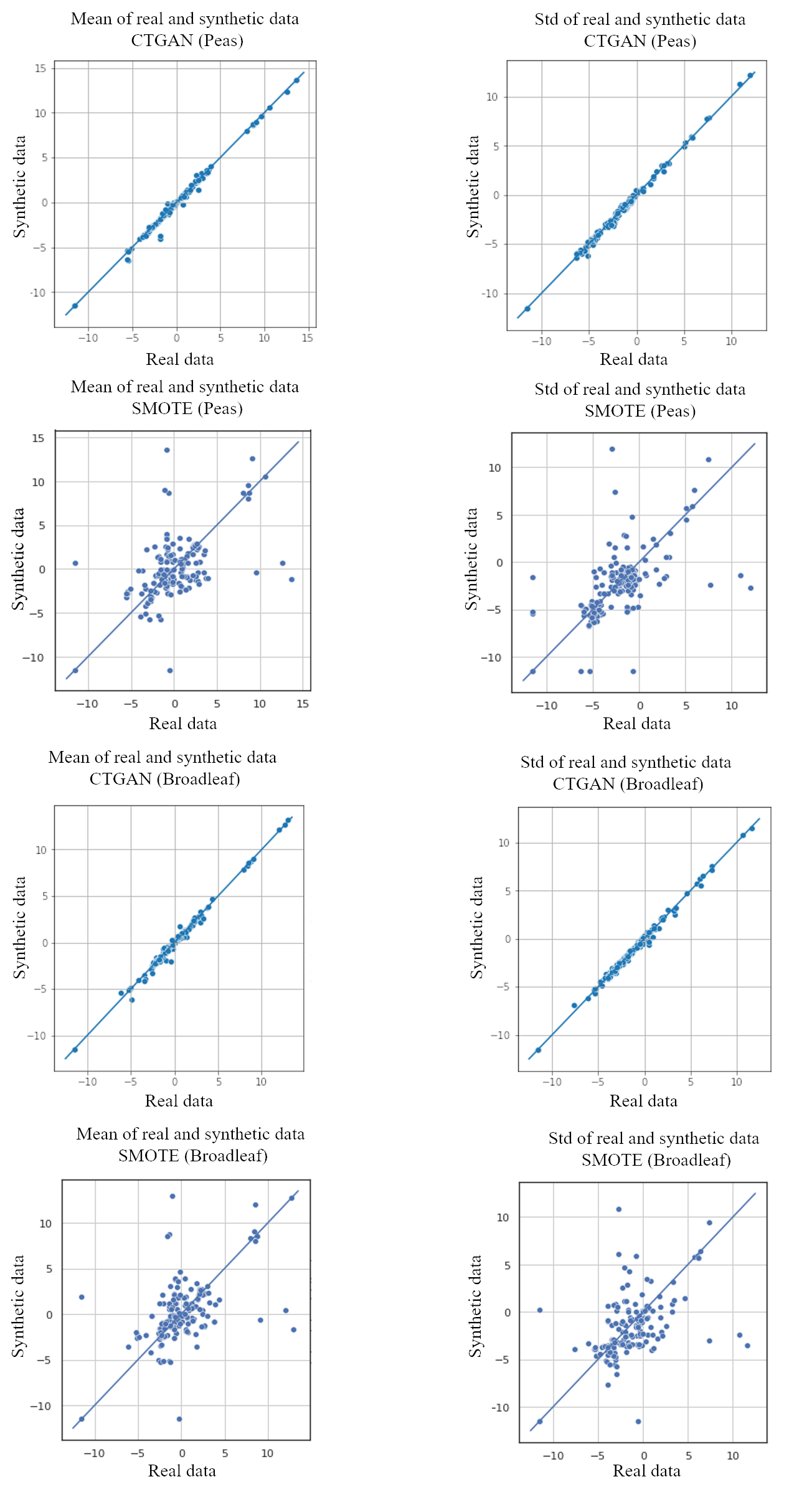}    
\caption{The correlation plots between means and standard deviations of synthetic data and real data. Each blue point corresponds to a feature extracted from SAR and Optical images.} 
\label{fig5}                                 
\end{center}                                 
\end{figure}

To further investigate the quality of the synthetically generated data, the means and standard deviations of real and synthetic data for the 168 optical and polarimetric features introduced in Section \ref{method} generated by CTGAN and SMOTE for the Peas and Broadleaf classes. Figure \ref{fig5} displays the correlation plots between means and standard deviations of real and synthetic data. The position of each scatter point is the mean (or standard deviation) of the real data versus the synthetic data for one employed feature. The plot in which the positions of points are closer to the line of equivalence (Y=X line) implies that the synthetic data have similar statistical properties with the real data, which reflects more similarity between distributions. The correlation plots show that the means and standard deviations of the features generated by CTGAN have a higher correlation to the real data compared to those generated by SMOTE. This indicates that CTGAN is able to accurately reproduce the distribution of the real data compared to SMOTE.

\begin{figure*}[ht!]
\begin{center}
\includegraphics[width=0.8\textwidth]{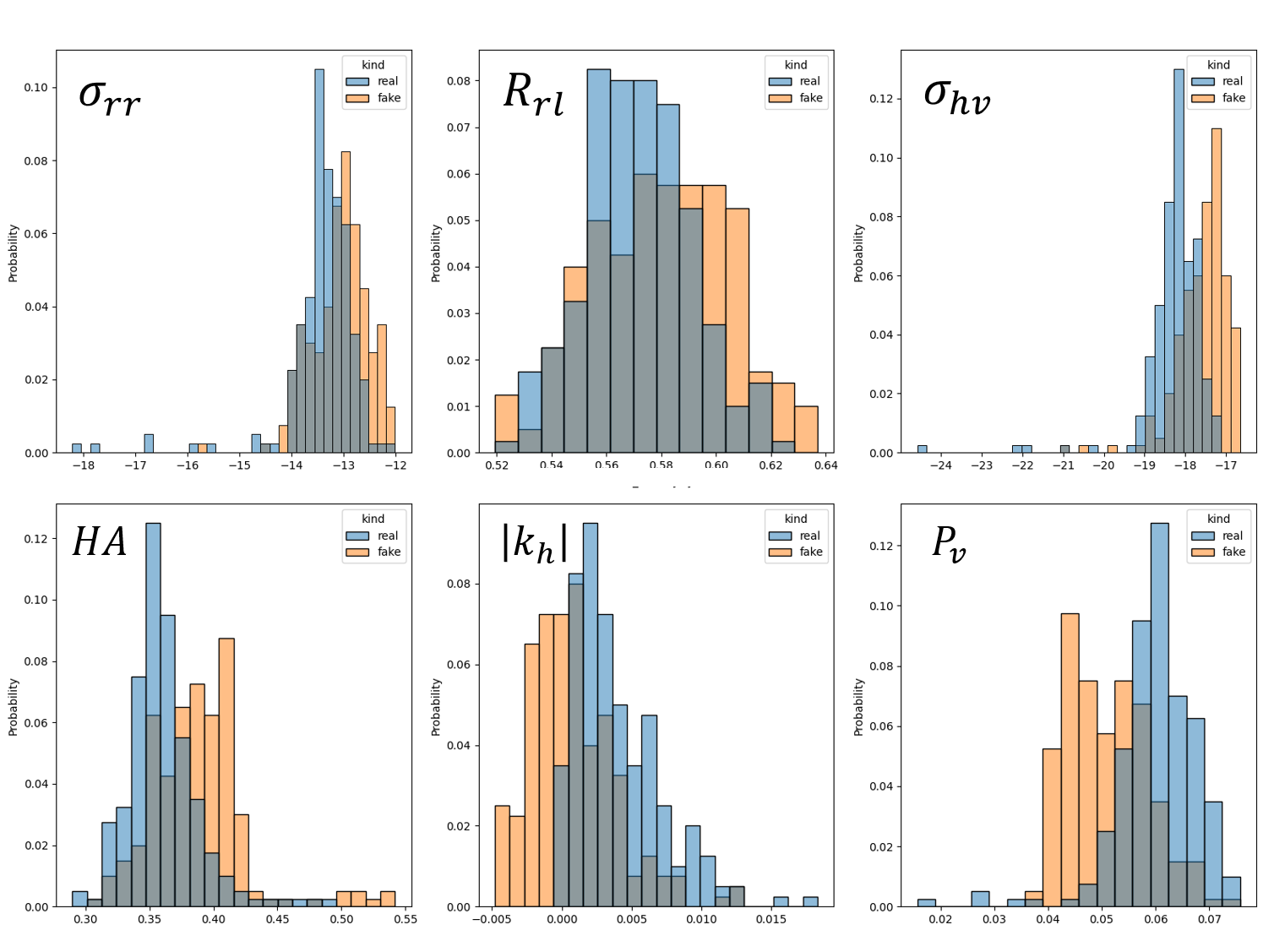} 
\caption{The data distribution of 6 exemplary features generated by CTGAN versus the distribution of the real data for the Peas class.}
\label{fig11}                                 
\end{center}                                 
\end{figure*}

\begin{figure*}[ht!]
\begin{center}
\includegraphics[width=0.8\textwidth]{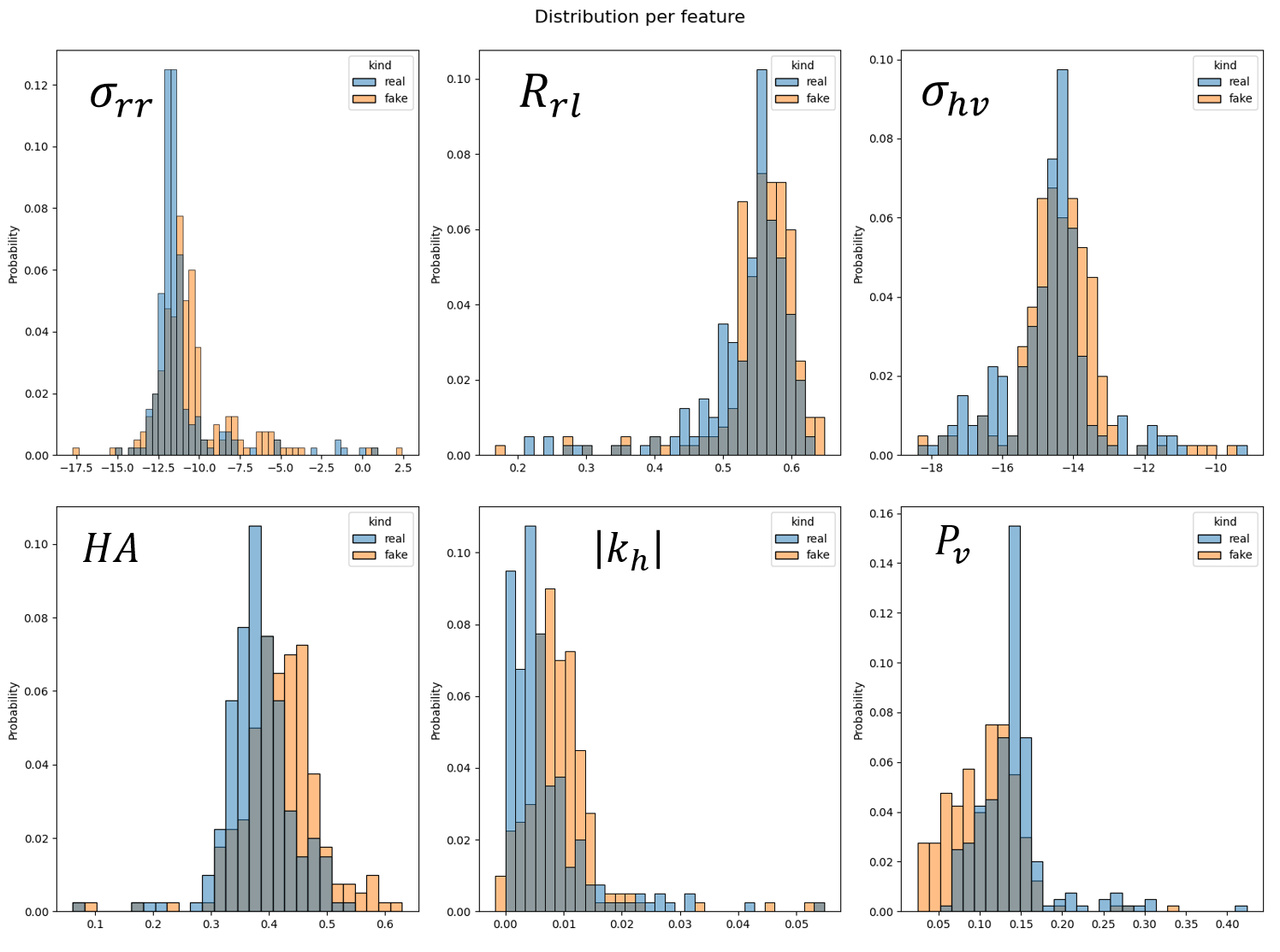}    
\caption{The data distribution of 6 exemplary features generated by CTGAN versus distribution of the real data for Broadleaf class.} 
\label{fig12}                                 
\end{center}                                 
\end{figure*}

Figures \ref{fig11} and \ref{fig12} provide a detailed comparison of feature distributions between real and synthetic data generated by CTGAN for two minority classes. Figure \ref{fig11} focuses on the Peas class, showing the distributions for 6 exemplary features in real data (blue columns) versus synthetic data (orange columns). Figure \ref{fig12} repeats this comparison for those 6 features of the Broadleaf class. In both figures, the synthetic data distributions generated by CTGAN closely match those of the corresponding real data features. This consistency demonstrates CTGAN's ability to accurately model the underlying distributions existing in the real data. Notably, CTGAN is also capable of reconstructing synthetic data in a wider range compared to real features. For instance the distribution of, some synthetic features in Figures \ref{fig11} and \ref{fig12} extend beyond the maximum and minimum values of the real data. This extension reduces the risk of overfitting during subsequent classifier training, as the models are exposed to a more diverse representation of each feature data during training. Overall, these distribution comparisons provide strong evidence that CTGAN can successfully capture the statistical properties of the real data, thereby can generate synthetic data well representative of the original samples. This fidelity facilitates the effective application of synthetic data for classification tasks.


\section{Conclusion}
This article investigated the performance of the CTGAN model to generate synthetic data to reduce the impact of insufficient samples in crop classification. To study this issue, the features extracted from the optical and SAR images obtained from the RapidEye and UASAR sensors were employed. In this research, by using three classifiers XGBoost, RF, and KNN, the quality of synthetic data generated by CTGAN (as a state-of-the-art method) was evaluated in comparison to RUS, ROS, and SMOTE. In general, the results of the research demonstrated the significant superiority of the CTGAN network over the comparative algorithms. While SMOTE generated synthetic data without considering the distribution of real data, the CTGAN method took account of data distribution during the data generation. Furthermore, RUS and ROS did not generate desirable data to considerably improve the performance of classifiers compared to the CTGAN model. Also, the quality of the synthetic data generated by CTGAN was evaluated by comparing it to real data using statistical metrics. Specifically, the mean, standard deviation, and distributions of different features were measured and compared. The results showed that the data produced by CTGAN exhibited similar to the real data across the aforementioned statistical metrics. This indicates that the synthetic data generated by CTGAN accurately reflects the real data distribution. Therefore, the CTGAN network is a better alternative to the basic methods of generating synthetic datasets.\textcolor{blue}{ However, it is important to acknowledge that using CTGAN for synthetic data generation has certain limitations. One such limitation is the requirement of a minimum amount of data for training. CTGAN relies on a sufficient quantity of training data to effectively learn the underlying data distribution and capture the intricate dependencies within the dataset. 
Additionally, it is worth noting that the training process of CTGAN can be more time-consuming compared to traditional data generation methods. CTGAN involves training a generative model that learns the complex patterns and relationships inherent in the data. This training process typically requires multiple iterations and can be computationally intensive, especially when dealing with large and high-dimensional datasets. The processing time required for training CTGAN should be considered when deciding on the appropriate data generation approach, especially in time-sensitive applications or scenarios with resource constraints.
Despite these limitations, the benefits of CTGAN should not be overlooked. CTGAN excels at capturing the underlying data distribution and generating synthetic samples that closely resemble the real data. It has the potential to overcome the limitations of traditional methods by preserving complex relationships and dependencies present in the original dataset. Additionally, CTGAN offers more flexibility in generating synthetic data with desired characteristics, allowing researchers to control specific features or adjust the balance between classes.}

\textcolor{blue}{Some important directions for future work include: 1) Applying CTGAN and comparing its performance to other generative models in other remote sensing applications beyond crop classification. This could include tasks like land cover mapping, object detection in aerial/satellite imagery, and environmental monitoring. Evaluating generative solutions across different problem types would expand our understanding of their capabilities and limitations. 2) Leveraging synthetic data generation to address lack of training samples in various remote sensing and geospatial problems beyond agriculture, such as damage assessment from natural disasters, urban development monitoring, infrastructure mapping, and species habitat modeling, where limited labeled data exists, generative models may help boost predictive accuracy. 3) Developing new generative model architectures and training procedures specialized for different remote sensing inputs. }

\section*{Acknowledgments}
The authors would like to present their acknowledgments to the JPL NASA, MacDonald, Dettwiler and Associates Ltd, the German Aerospace Center (DLR) DLR, the SMAPVEX 2012 team, the Agriculture and Agri-Food Canada, and Dr. Saeid Homayouni, from the Dept. of Geography, Environment, and Geomatics of the University of Ottawa, Canada, for providing the PolSAR and the fields survey data used in this research.

\bibliography{reference.bib}
\end{document}